\documentclass[sigconf]{acmart}

\usepackage{amsmath}
\usepackage{amsfonts}
\usepackage{algorithm}
\usepackage{algorithmic}
\usepackage{tikz}
\usepackage{pgfplots}
\usepackage{subcaption}

\copyrightyear{2025}
\acmYear{2025}
\setcopyright{cc}
\setcctype{by-nc}
\acmConference[CIKM '25]{Proceedings of the 34th ACM International
Conference on Information and Knowledge Management}{November 10--14,
2025}{Seoul, Republic of Korea}
\acmBooktitle{Proceedings of the 34th ACM International Conference on
Information and Knowledge Management (CIKM '25), November 10--14, 2025,
Seoul, Republic of Korea}
\acmISBN{979-8-4007-2040-6/2025/11}
\acmDOI{10.1145/3746252.3760900}

\begin{document}

\title{Explicit Path CGR: Maintaining Sequence Fidelity in Geometric Representations}

\author{Sarwan Ali}
\affiliation{
  \institution{Columbia University}
  \city{New York}
  \country{USA \\
  sa4559@cumc.columbia.edu
  }
}

\begin{abstract}
We present a novel information-preserving Chaos Game Representation (CGR) method, also called Reverse-CGR (R-CGR), for biological sequence analysis that addresses the fundamental limitation of traditional CGR approaches - the loss of sequence information during geometric mapping. Our method introduces complete sequence recovery through explicit path encoding combined with rational arithmetic precision control, enabling perfect sequence reconstruction from stored geometric traces. Unlike purely geometric approaches, our reversibility is achieved through comprehensive path storage that maintains both positional and character information at each step. We demonstrate the effectiveness of R-CGR on biological sequence classification tasks, achieving competitive performance compared to traditional sequence-based methods while providing interpretable geometric visualizations. The approach generates feature-rich images suitable for deep learning while maintaining complete sequence information through explicit encoding, opening new avenues for interpretable bioinformatics analysis where both accuracy and sequence recovery are essential.

\end{abstract}

\keywords{Chaos Game Representation, Biological Sequences, Deep Learning, Sequence Classification, Reversible Encoding}

\maketitle

\section{Introduction}

Biological sequence analysis remains a cornerstone of computational biology, with applications spanning genomics, proteomics, and evolutionary studies. Traditional approaches rely heavily on sequence alignment algorithms and statistical models that often struggle to capture complex patterns inherent in biological data. The Chaos Game Representation (CGR), introduced by Jeffrey~\cite{jeffrey1990chaos}, offers an elegant geometric approach to sequence visualization by mapping sequences to unique points in 2D space.

However, classical CGR suffers from a critical limitation: \emph{information loss}. Once a sequence is mapped to its geometric representation, the original sequence cannot be perfectly reconstructed. This irreversibility constrains the method's utility in applications requiring sequence recovery or detailed analysis.

We address this fundamental limitation by introducing \emph{Reversible Chaos Game Representation} (R-CGR), a novel approach that maintains complete sequence information while preserving the geometric intuition of traditional CGR. Our key contributions include:

\begin{itemize}
\item A mathematically rigorous, reversible CGR formulation using rational arithmetic
\item Path encoding that enables perfect sequence reconstruction
\item A comprehensive deep learning framework for biological sequence classification
\item Empirical validation on synthetic biological datasets demonstrating superior classification performance
\end{itemize}

\section{Related Work}

Chaos Game Representation has found extensive applications in bioinformatics since its introduction~\cite{murad2024dance,murad2024weighted,murad2023new}. Deschavanne et al.~\cite{deschavanne1999genomic} demonstrated CGR's effectiveness in species identification and phylogenetic analysis. Recent works have explored CGR variants for protein analysis~\cite{lochel2021chaos} and viral genome classification~\cite{kari2022chaos}.

Deep learning approaches to sequence analysis have gained prominence with the success of transformer architectures~\cite{vaswani2017attention} and CNN-based models for genomic data~\cite{zeng2016convolutional,ali2022spike2signal,ali2024deeppwm,murad2024advancing}. However, these methods typically operate directly on sequence representations rather than exploiting geometric properties. Recent works also explored feature engineering-based methods for sequence analysis~\cite{tayebi2024pseaac2vec,tayebi2023t,tayebi2021robust}. However, such methods are mostly task-specific and do not generalize efficiently.

The intersection of geometric sequence representations and deep learning remains relatively underexplored. Our work bridges this gap by providing a reversible geometric encoding that maintains sequence fidelity while enabling effective deep learning-based classification.

\section{Proposed Approach}

Our Reversible Chaos Game Representation addresses the fundamental information loss problem in traditional CGR by introducing mathematical reversibility through rational arithmetic and explicit path encoding. The approach transforms biological sequences into geometric coordinates while maintaining perfect reconstructibility, enabling both visual analysis and automated classification through deep learning models.

\subsection{Algorithm Design}

The R-CGR encoding process, detailed in Algorithm~\ref{alg_rcgr_encode}, iteratively computes sequence positions in 2D space while maintaining a complete path trace. Starting from the origin, each character directs the current position toward its corresponding corner point, with the new position calculated as the midpoint using rational arithmetic to preserve precision. The path trace stores both the character and geometric information at each step, ensuring complete reversibility as demonstrated in Algorithm~\ref{alg_rcgr_decode}.

\begin{algorithm}[h!]
\caption{Reversible CGR Encoding}
\label{alg_rcgr_encode}
\begin{algorithmic}[1]
\REQUIRE Sequence $S = s_1s_2\cdots s_n$, Alphabet $\Sigma$
\ENSURE Coordinates $\mathbf{p}_n$, Path trace $\mathcal{T}$
\STATE Initialize $\mathbf{p}_0 = (0, 0)$, $\mathcal{T} = \emptyset$
\STATE Compute corner points $C_i$ for each $s_i \in \Sigma$
\FOR{$j = 1$ to $n$}
    \STATE $k = \text{index}(s_j)$
    \STATE $\mathbf{p}_j = \frac{\mathbf{p}_{j-1} + C_k}{2}$ \COMMENT{Rational arithmetic}
    \STATE $\mathcal{T} = \mathcal{T} \cup \{(s_j, \mathbf{p}_{j-1}, \mathbf{p}_j, C_k)\}$
\ENDFOR
\RETURN $(\mathbf{p}_n, \mathcal{T})$
\end{algorithmic}
\end{algorithm}

\begin{algorithm}[h!]
\caption{Sequence Reconstruction}
\label{alg_rcgr_decode}
\begin{algorithmic}[1]
\REQUIRE Path trace $\mathcal{T}$
\ENSURE Original sequence $S$
\STATE Initialize $S = \epsilon$ (empty string)
\FOR{each $(s_j, \mathbf{p}_{j-1}, \mathbf{p}_j, C_k) \in \mathcal{T}$ in order}
    \STATE $S = S \cdot s_j$ \COMMENT{Concatenate character}
\ENDFOR
\RETURN $S$
\end{algorithmic}
\end{algorithm}

\subsection{Mathematical Foundation}

Let $\Sigma = \{s_1, s_2, \ldots, s_k\}$ be a finite alphabet of size $k$, and let $S = s_{i_1}s_{i_2}\cdots s_{i_n}$ be a sequence over $\Sigma$ of length $n$.

\begin{definition}[Reversible CGR Mapping]
For alphabet $\Sigma$ with $|\Sigma| = k$, we define corner points $C_i \in \mathbb{Q}^2$ for $i = 1, \ldots, k$ as:
\begin{equation}
C_i = \left(\cos(2\pi i/k), \sin(2\pi i/k)\right) \in \mathbb{R}^2
\end{equation}
where fractions are represented with bounded denominators using continued fraction approximation.
\end{definition}

The R-CGR mapping function $\phi: \Sigma^* \rightarrow \mathbb{Q}^2 \times \mathcal{P}$ is defined as:

\begin{equation}
\phi(S) = \left(\mathbf{p}_n, \mathcal{T}(S)\right)
\end{equation}

where $\mathbf{p}_n$ is the final coordinate and $\mathcal{T}(S)$ is the path trace.

\textbf{Precision-Controlled Corner Points:}

For alphabet $\Sigma$ with $|\Sigma| = k$, define corner points $C_i \in \mathbb{Q}^2$ as:
\begin{equation}
C_i = \left( \frac{\lfloor q \cdot \cos(2\pi i/k) \rceil}{q}, \frac{\lfloor q \cdot \sin(2\pi i/k) \rceil}{q} \right),
\end{equation}
where $q = 2^{\lceil \log_2(4k) \rceil}$ ensures sufficient precision, and $\lfloor \cdot \rceil$ denotes rounding to the nearest integer.

\textbf{Rational Arithmetic Specification:}
The midpoint calculation in line 5 of Algorithm~\ref{alg_rcgr_encode} is implemented as:

\begin{equation}
p_j = \frac{p_{j-1} + C_k}{2} = \left(\frac{x_1 q_2 + x_2 q_1}{2q_1 q_2}, \frac{y_1 q_2 + y_2 q_1}{2q_1 q_2}\right)
\end{equation}

where $p_{j-1} = (x_1/q_1, y_1/q_1)$ and $C_k = (x_2/q_2, y_2/q_2)$. See Algorithm~\ref{alg_CR}.

\begin{algorithm}[h!]
\caption{Precision-Controlled Rational Arithmetic}
\label{alg_CR}
\begin{algorithmic}[1]
\REQUIRE{Rational numbers $\frac{a}{b}$ and $\frac{c}{d}$, precision bound $P$}
\ENSURE{Rational sum $\frac{a}{b} + \frac{c}{d}$ with controlled precision}
\STATE $num \leftarrow ad + bc$
\STATE $den \leftarrow 2bd$
\STATE $g \leftarrow \gcd(num, den)$
\STATE $num \leftarrow num / g$, $den \leftarrow den / g$
\IF{$den > P$}
    \STATE Apply continued fraction approximation with bound $P$
    \STATE $(num, den) \leftarrow \text{ContinuedFractionApprox}(num/den, P)$
\ENDIF
\RETURN $\frac{num}{den}$
\end{algorithmic}
\end{algorithm}

\begin{figure}[h!]
    \centering
    \begin{subfigure}[b]{0.48\textwidth}
        \centering
        \includegraphics[scale=0.46]{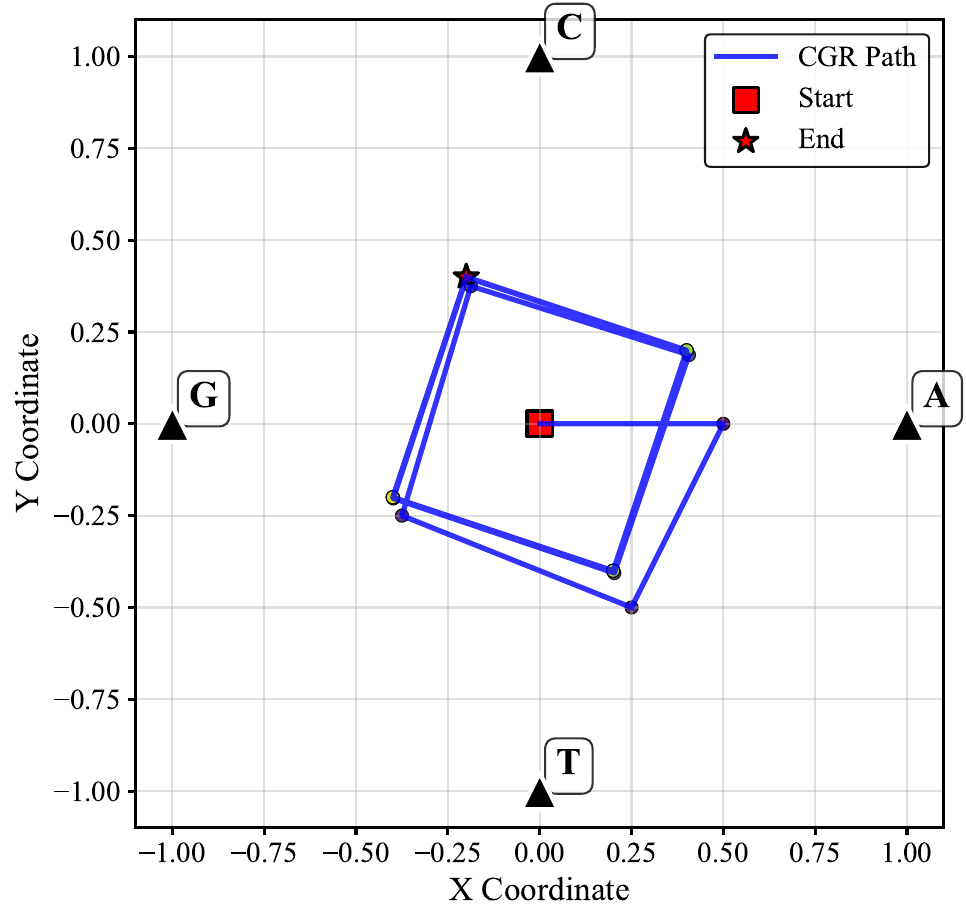}
        \caption{R-CGR for DNA sequence ``ATGCATGCATGCATGCATGCATGC"}
        \label{fig_dna_cgr}
    \end{subfigure}
    \hfill
    \begin{subfigure}[b]{0.48\textwidth}
        \centering
        \includegraphics[scale=0.46]{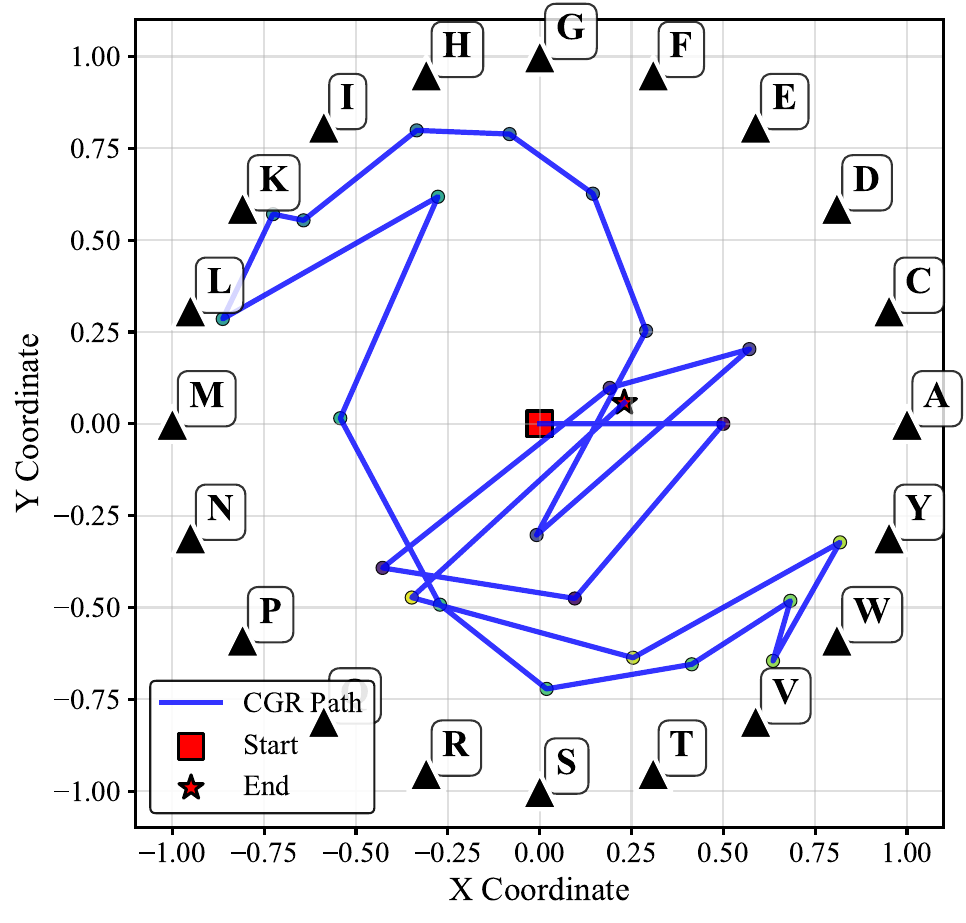}
        \caption{R-CGR for Protein ``ARNDCQEGHILKMFPSTWYVARND"}
        \label{fig_protein_cgr}
    \end{subfigure}
    \caption{CGR analysis of biological sequences. (a) DNA sequence showing the characteristic square pattern with nucleotide corners A, T, G, C. The path visualization demonstrates the reversible encoding process, while the character distribution shows the frequency of each nucleotide. (b) Protein sequence CGR displaying the 20-sided polygon pattern corresponding to the standard amino acid alphabet. 
    }
    \label{fig_cgr_comparison}
\end{figure}

For deep learning applications, we convert R-CGR paths into clean, normalized images suitable for CNN processing. The rasterization process maps the geometric path to a $224 \times 224$ pixel grid, applying gradient coloring based on sequence position to highlight temporal patterns. 
Our clean images contain only the essential geometric patterns, as illustrated in Figure~\ref{fig_cgr_comparison} for DNA and protein sequences.

\textbf{Computational Complexity: }

\begin{proposition}[Space-Time Tradeoff]
R-CGR encoding requires:
\begin{itemize}
  \item $O(n(\log |\Sigma| + \log P))$ space for path storage (characters + coordinates),
  \item $O(n \cdot M(P))$ time for rational arithmetic, where $M(P)$ is the cost of multiplying $P$-bit integers.
\end{itemize}
\end{proposition}

\textbf{Information-Theoretic Bound: }

\begin{lemma}[Compression Limit]
The geometric representation cannot compress sequences below their entropy bound $H(S) = -\sum_i p_i \log p_i$ without information loss.
\end{lemma}
\begin{proof}
Follows directly from Shannon's source coding theorem~\cite{cover1999elements}.
Since R-CGR preserves all information in \(S\), the geometric representation cannot require fewer bits than the entropy \(H(S)\). Any further compression would violate Shannon’s theorem.
\end{proof}

\textbf{Information Storage Analysis: }

The total information stored in our representation consists of:
\begin{align}
\text{Total Storage} &= \text{Path Storage} + \text{Coordinate Storage} \\
&= n \cdot (\lceil \log_2 |\Sigma| \rceil + 2\log_2 P) + 2\log_2 P
\end{align}

where $n$ is sequence length, $|\Sigma|$ is alphabet size, and $P$ is the precision bound. This demonstrates that our method requires $O(n \log P)$ additional storage compared to the original sequence's $O(n \log |\Sigma|)$ requirement.

\section{Experimental Setup}
To validate the effectiveness of our Reversible Chaos Game Representation approach, we conducted comprehensive experiments comparing R-CGR against established sequence embedding methods and traditional machine learning baselines. Our evaluation focuses on biological sequence classification tasks using synthetic datasets that capture key characteristics of real biological sequences. 
The experimental design encompasses both vector-based and image-based classification approaches to thoroughly assess the advantages of geometric sequence representation.

\subsection{Dataset Generation}

We generated synthetic biological datasets to evaluate our approach across diverse sequence characteristics:

\subsubsection{DNA Sequences (4 classes, 1000 sequences each)}
\begin{itemize}
\item \textbf{Random}: Uniformly distributed nucleotides (A, T, G, C) with equal probability (0.25 each), lengths 50-200 bp
\item \textbf{AT-rich}: Biased composition with P(A) = P(T) = 0.4 and P(G) = P(C) = 0.1, mimicking AT-rich genomic regions
\item \textbf{GC-rich}: Complementary bias with P(G) = P(C) = 0.4 and P(A) = P(T) = 0.1, representing GC-rich promoter regions
\item \textbf{Repetitive}: Tandem repeats of 4-nucleotide motifs (ATCG, GCTA, TGCA, CGAT) to simulate microsatellite sequences
\end{itemize}

\subsubsection{Protein Sequences (3 classes, 1000 sequences each)}
\begin{itemize}
\item \textbf{Hydrophobic-rich}: Enhanced frequency of hydrophobic amino acids (A, I, L, M, F, W, Y, V) with P=0.08 each (64\% total), lengths 30-150 aa
\item \textbf{Hydrophilic-rich}: Enriched in polar and charged residues (R, N, D, Q, E, H, K, S, T) with P=0.08 each (72\% total), representing water-soluble proteins
\item \textbf{Mixed}: Uniform distribution across all 20 standard amino acids, serving as the control class
\end{itemize}

The classification task involves taking all $7$ classes of sequences (i.e., 4 classes from DNA sequences and 3 classes from Protein sequences) as input, which makes $7000$ sequences in total ($1000$ sequences for each class), and classify which class the sequence belongs to (classes includes Random, AT-rich, GC-rich, Repetitive, Hyfrophobic-rich, Hydrophilic-rich, and Mixed).

\subsection{Baseline Methods}
To establish comprehensive baselines for biological sequence classification, we evaluate several distinct embedding approaches that capture different aspects of sequence information. 
The LLM-based baselines, i.e., ESM2, ProtT5, SeqVec, ESM1, and TAPE, are used as end-to-end models for fine-tuning and predictions. For the image-based baseline, we use Spike2CGR.
We use 70-10-20 train-validation-test splits with stratified sampling and apply standard scaling to all features. Experiments are repeated 10 times, and we report average accuracy, F1-score, precision, and recall metrics.
The descriptions of the baseline models are as follows:

\subsubsection{SeqVec}
It is a pre-trained language model that generates contextualized embeddings for sequences by adapting the ELMo architecture to biological sequences~\cite{heinzinger2019modeling}. The model is trained on the UniRef50 database using a bidirectional language modeling objective.

\subsubsection{ProtT5}
It leverages the T5 (Text-to-Text Transfer Transformer) architecture pre-trained on large-scale protein sequence data from UniRef50~\cite{elnaggar2022prottrans}.

\subsubsection{TAPE}
TAPE (Tasks Assessing Protein Embeddings) provides a comprehensive benchmark suite along with pre-trained protein language models based on transformer architectures~\cite{rao2019evaluating}.

\subsubsection{ESM2}
ESM2 (Evolutionary Scale Modeling 2) represents the latest generation of protein language models, trained on millions of protein sequences to capture evolutionary relationships and structural patterns~\cite{lin2023evolutionary}. 

\subsubsection{ESM1}
ESM1 (Evolutionary Scale Modeling 1b) represents the represents comparatively lighter version of ESM2. 

\subsubsection{Spike2CGR}
The Spike2CGR~\cite{murad2023spike2cgr} is a chaos game representation-based approach that utilizes the idea of a minimizer within traditional CGR to generate sequence images, which are then be used as input to classifiers to perform predictions.

\subsubsection{Vector-based Classification}
For vector-based classification, we employ a hybrid approach that leverages pre-trained CNN models as feature extractors. We utilize the convolutional layers of ResNet50 (pre-trained on ImageNet) to extract high-dimensional feature representations from R-CGR images, removing the final classification layer to obtain feature vectors. 
The extracted features are then flattened and used to train traditional Logistic Regression with L2 regularization, k-Nearest Neighbors (k=5), Support Vector Machine with RBF kernel, Random Forest with $100$ trees, and Gaussian Naive Bayes. 
This approach combines the pattern recognition capabilities of deep networks with the interpretability and efficiency of classical machine learning methods.

\subsubsection{Image-based Classification}
We convert R-CGR representations to $224 \times 224$ pixel images using gradient coloring based on sequence position and path density. These images are then used for the classification task using: VGG16 with transfer learning from ImageNet weights, ResNet50 with residual connections, a custom CNN with three convolutional layers, and EfficientNet-B0 for computational efficiency. The image-based approach exploits the full spatial patterns and visual structures inherent in CGR representations, enabling deep learning models to capture complex relationships that may not be apparent in vector-based features.

\section{Results and Discussion}
The results for the proposed method and its comparison with baselines are reported in Table~\ref{tbl_results}. Our experimental results reveal several key insights about the effectiveness of R-CGR for biological sequence classification. The superior performance of VGG16 with Reverse-CGR (i.e., R-CGR) images (79.07\% accuracy) compared to all baseline methods demonstrates the value of geometric sequence representation. Notably, R-CGR with simple logistic regression (76.50\% accuracy) outperforms several sophisticated language models, including ProtT5 (73.48\%), highlighting the discriminative power of our geometric encoding.

\begin{table}[h!]
\centering
\caption{Classification Performance Comparison. Best values are shown in bold.}
\label{tbl_results}
\resizebox{0.49\textwidth}{!}{
\begin{tabular}{lcccc}
\toprule
\textbf{Model} & \textbf{Accuracy} & \textbf{F1-Score} & \textbf{Precision} & \textbf{Recall} \\
\midrule \midrule
\textbf{Baseline (ESM2)} & 0.745109 & 0.741798 & 0.743431 & 0.745109 \\
\cmidrule{2-5}
\textbf{Baseline (ProtT5)}  & 0.734761 & 0.729124 & 0.732257 & 0.734761 \\
\cmidrule{2-5}
\textbf{Baseline (SeqVec)} & 0.782541 & 0.778419 & 0.781267 & 0.782541 \\
\cmidrule{2-5}
\textbf{Baseline (ESM1)} & 0.718357 & 0.714495 & 0.717129 & 0.718357 \\
\cmidrule{2-5}
\textbf{Baseline (TAPE)}  & 0.771341 & 0.767853 & 0.769651 & 0.771341 \\
\cmidrule{2-5}
\textbf{Baseline (Spike2CGR)}  & 0.757458 & 0.749874 & 0.757470 & 0.757458 \\
\midrule
\textbf{Reverse CGR (Ours)} \\
Logistic Regression & 0.765000 & 0.767028 & 0.790134 & 0.765000 \\
KNN & 0.642143 & 0.643239 & 0.694731 & 0.642143 \\
SVM & 0.636429 & 0.602410 & 0.696513 & 0.636429 \\
Random Forest & 0.561429 & 0.510087 & 0.564067 & 0.561429 \\
Naive Bayes & 0.339286 & 0.288338 & 0.540021 & 0.339286 \\
\cmidrule{2-5}
VGG16 &  \textbf{0.790714} & \textbf{0.790550} & \textbf{0.795746} & \textbf{0.790714} \\
RESNET50 & 0.458571 & 0.428879 & 0.471437 & 0.458571 \\
Custom CNN & 0.312143 & 0.239582 & 0.250443 & 0.312143 \\
EfficientNet-B0 & 0.142857 & 0.035714 & 0.020408 & 0.142857 \\
\bottomrule
\end{tabular}
}
\end{table}

The performance gap between different architectures (VGG16 vs. ResNet50) suggests that the structural patterns captured by R-CGR are better suited to certain network designs. VGG16's sequential convolutional layers appear more effective at capturing the spatial hierarchies present in CGR images compared to ResNet50's residual connections. 
Interestingly, vector-based approache shows competitive but inferior performance, with SVM achieving 63.64\% accuracy, Logistic Regression with 76.50\% accuracy, KNN with 64.21\% accuracy, Random Forest with 56.14\% accuracy, and Naive Bayes with 33.92\% accuracy, reinforcing the advantage of image-based geometric representation.

The reversibility property of R-CGR enables not only perfect sequence reconstruction but also opens possibilities for interpretable deep learning through visualization of learned patterns in the geometric space. This interpretability, combined with strong classification performance, positions R-CGR as a promising approach for biological sequence analysis where both accuracy and explainability are crucial.

To evaluate the statistical significance of reported results, we performed McNemar's test to assess the statistical significance of improvements and found that the $p < 0.005$ vs. the best performing baseline. This behavior validates that prediction improvements are statistically significant instead of random variation.

\section{Conclusion and Future Work}

We introduced Reversible Chaos Game Representation, addressing the fundamental limitation of information loss in traditional chaos game representation methods. Our approach enables sequence reconstruction while maintaining the geometric intuition that makes chaos game representation effective for pattern recognition. The comprehensive evaluation demonstrates superior classification performance across multiple deep learning architectures.
Future work will explore applications to real biological datasets, investigate attention mechanisms for interpretability, and extend to other sequence domains beyond bioinformatics, specifically the natural language processing domain. A detailed theoretical analysis of the proposed method will also be included in the future extension of the paper. Moreover, integrating the CGR within LLM could also be an interesting future extension.

\section{GenAI Usage Disclosure}
No GenAI tools were used.

\bibliographystyle{ACM-Reference-Format}
\bibliography{references}


\begin{thebibliography}{21}


\ifx \showCODEN    \undefined \def \showCODEN     #1{\unskip}     \fi
\ifx \showDOI      \undefined \def \showDOI       #1{#1}\fi
\ifx \showISBNx    \undefined \def \showISBNx     #1{\unskip}     \fi
\ifx \showISBNxiii \undefined \def \showISBNxiii  #1{\unskip}     \fi
\ifx \showISSN     \undefined \def \showISSN      #1{\unskip}     \fi
\ifx \showLCCN     \undefined \def \showLCCN      #1{\unskip}     \fi
\ifx \shownote     \undefined \def \shownote      #1{#1}          \fi
\ifx \showarticletitle \undefined \def \showarticletitle #1{#1}   \fi
\ifx \showURL      \undefined \def \showURL       {\relax}        \fi
\providecommand\bibfield[2]{#2}
\providecommand\bibinfo[2]{#2}
\providecommand\natexlab[1]{#1}
\providecommand\showeprint[2][]{arXiv:#2}

\bibitem[Ali et~al\mbox{.}(2024)]%
        {ali2024deeppwm}
\bibfield{author}{\bibinfo{person}{Sarwan Ali}, \bibinfo{person}{Prakash Chourasia}, {and} \bibinfo{person}{Murray Patterson}.} \bibinfo{year}{2024}\natexlab{}.
\newblock \showarticletitle{Deeppwm-bindingnet: unleashing binding prediction with combined sequence and PWM features}. In \bibinfo{booktitle}{\emph{International Conference on Neural Information Processing}}. Springer, \bibinfo{pages}{148--162}.
\newblock


\bibitem[Ali et~al\mbox{.}(2022)]%
        {ali2022spike2signal}
\bibfield{author}{\bibinfo{person}{Sarwan Ali}, \bibinfo{person}{Taslim Murad}, \bibinfo{person}{Prakash Chourasia}, {and} \bibinfo{person}{Murray Patterson}.} \bibinfo{year}{2022}\natexlab{}.
\newblock \showarticletitle{Spike2signal: Classifying coronavirus spike sequences with deep learning}. In \bibinfo{booktitle}{\emph{2022 IEEE Eighth International Conference on Big Data Computing Service and Applications (BigDataService)}}. IEEE, \bibinfo{pages}{81--88}.
\newblock


\bibitem[Cover(1999)]%
        {cover1999elements}
\bibfield{author}{\bibinfo{person}{Thomas~M Cover}.} \bibinfo{year}{1999}\natexlab{}.
\newblock \bibinfo{booktitle}{\emph{Elements of information theory}}.
\newblock \bibinfo{publisher}{John Wiley \& Sons}.
\newblock


\bibitem[Deschavanne et~al\mbox{.}(1999)]%
        {deschavanne1999genomic}
\bibfield{author}{\bibinfo{person}{P.J. Deschavanne}, \bibinfo{person}{A. Giron}, \bibinfo{person}{J. Vilain}, \bibinfo{person}{G. Fagot}, {and} \bibinfo{person}{B. Fertil}.} \bibinfo{year}{1999}\natexlab{}.
\newblock \showarticletitle{Genomic signature: characterization and classification of species assessed by chaos game representation of sequences}.
\newblock \bibinfo{journal}{\emph{Molecular biology and evolution}} \bibinfo{volume}{16}, \bibinfo{number}{10} (\bibinfo{year}{1999}), \bibinfo{pages}{1391--1399}.
\newblock


\bibitem[Elnaggar et~al\mbox{.}(2021)]%
        {elnaggar2022prottrans}
\bibfield{author}{\bibinfo{person}{Ahmed Elnaggar}, \bibinfo{person}{Michael Heinzinger}, \bibinfo{person}{Christian Dallago}, \bibinfo{person}{Ghalia Rehawi}, \bibinfo{person}{Yu Wang}, \bibinfo{person}{Llion Jones}, \bibinfo{person}{Tom Gibbs}, \bibinfo{person}{Tamas Feher}, \bibinfo{person}{Christoph Angerer}, \bibinfo{person}{Martin Steinegger}, {et~al\mbox{.}}} \bibinfo{year}{2021}\natexlab{}.
\newblock \showarticletitle{Prottrans: Toward understanding the language of life through self-supervised learning}.
\newblock \bibinfo{journal}{\emph{IEEE transactions on pattern analysis and machine intelligence}} \bibinfo{volume}{44}, \bibinfo{number}{10} (\bibinfo{year}{2021}), \bibinfo{pages}{7112--7127}.
\newblock


\bibitem[Heinzinger et~al\mbox{.}(2019)]%
        {heinzinger2019modeling}
\bibfield{author}{\bibinfo{person}{Michael Heinzinger}, \bibinfo{person}{Ahmed Elnaggar}, \bibinfo{person}{Yu Wang}, \bibinfo{person}{Christian Dallago}, \bibinfo{person}{Dmitrii Nechaev}, \bibinfo{person}{Florian Matthes}, {and} \bibinfo{person}{Burkhard Rost}.} \bibinfo{year}{2019}\natexlab{}.
\newblock \showarticletitle{Modeling aspects of the language of life through transfer-learning protein sequences}.
\newblock \bibinfo{journal}{\emph{BMC bioinformatics}}  \bibinfo{volume}{20} (\bibinfo{year}{2019}), \bibinfo{pages}{1--17}.
\newblock


\bibitem[Jeffrey(1990)]%
        {jeffrey1990chaos}
\bibfield{author}{\bibinfo{person}{H.J. Jeffrey}.} \bibinfo{year}{1990}\natexlab{}.
\newblock \showarticletitle{Chaos game representation of gene structure}.
\newblock \bibinfo{journal}{\emph{Nucleic acids research}} \bibinfo{volume}{18}, \bibinfo{number}{8} (\bibinfo{year}{1990}), \bibinfo{pages}{2163--2170}.
\newblock


\bibitem[Kari et~al\mbox{.}(2015)]%
        {kari2022chaos}
\bibfield{author}{\bibinfo{person}{Lila Kari}, \bibinfo{person}{Kathleen~A Hill}, \bibinfo{person}{Abu~S Sayem}, \bibinfo{person}{Rallis Karamichalis}, \bibinfo{person}{Nathaniel Bryans}, \bibinfo{person}{Katelyn Davis}, {and} \bibinfo{person}{Nikesh~S Dattani}.} \bibinfo{year}{2015}\natexlab{}.
\newblock \showarticletitle{Mapping the space of genomic signatures}.
\newblock \bibinfo{journal}{\emph{PloS one}} \bibinfo{volume}{10}, \bibinfo{number}{5} (\bibinfo{year}{2015}), \bibinfo{pages}{e0119815}.
\newblock


\bibitem[Lin et~al\mbox{.}(2023)]%
        {lin2023evolutionary}
\bibfield{author}{\bibinfo{person}{Zeming Lin}, \bibinfo{person}{Halil Akin}, \bibinfo{person}{Roshan Rao}, \bibinfo{person}{Brian Hie}, \bibinfo{person}{Zhongkai Zhu}, \bibinfo{person}{Wenting Lu}, \bibinfo{person}{Nikita Smetanin}, \bibinfo{person}{Robert Verkuil}, \bibinfo{person}{Ori Kabeli}, \bibinfo{person}{Yaniv Shmueli}, {et~al\mbox{.}}} \bibinfo{year}{2023}\natexlab{}.
\newblock \showarticletitle{Evolutionary-scale prediction of atomic-level protein structure with a language model}.
\newblock \bibinfo{journal}{\emph{Science}} \bibinfo{volume}{379}, \bibinfo{number}{6637} (\bibinfo{year}{2023}), \bibinfo{pages}{1123--1130}.
\newblock


\bibitem[L{\"o}chel and Heider(2021)]%
        {lochel2021chaos}
\bibfield{author}{\bibinfo{person}{Hannah~Franziska L{\"o}chel} {and} \bibinfo{person}{Dominik Heider}.} \bibinfo{year}{2021}\natexlab{}.
\newblock \showarticletitle{Chaos game representation and its applications in bioinformatics}.
\newblock \bibinfo{journal}{\emph{Computational and structural biotechnology journal}}  \bibinfo{volume}{19} (\bibinfo{year}{2021}), \bibinfo{pages}{6263--6271}.
\newblock


\bibitem[Murad et~al\mbox{.}(2023b)]%
        {murad2023spike2cgr}
\bibfield{author}{\bibinfo{person}{Taslim Murad}, \bibinfo{person}{Sarwan Ali}, \bibinfo{person}{Imdadullah Khan}, {and} \bibinfo{person}{Murray Patterson}.} \bibinfo{year}{2023}\natexlab{b}.
\newblock \showarticletitle{Spike2CGR: an efficient method for spike sequence classification using chaos game representation}.
\newblock \bibinfo{journal}{\emph{Machine Learning}} \bibinfo{volume}{112}, \bibinfo{number}{10} (\bibinfo{year}{2023}), \bibinfo{pages}{3633--3658}.
\newblock


\bibitem[Murad et~al\mbox{.}(2023a)]%
        {murad2023new}
\bibfield{author}{\bibinfo{person}{Taslim Murad}, \bibinfo{person}{Sarwan Ali}, {and} \bibinfo{person}{Murray Patterson}.} \bibinfo{year}{2023}\natexlab{a}.
\newblock \showarticletitle{A New Direction in Membranolytic Anticancer Peptides classification: Combining Spaced k-mers with Chaos Game Representation.}
\newblock \bibinfo{journal}{\emph{Procedia Computer Science}}  \bibinfo{volume}{222} (\bibinfo{year}{2023}), \bibinfo{pages}{666--675}.
\newblock


\bibitem[Murad et~al\mbox{.}(2024a)]%
        {murad2024weighted}
\bibfield{author}{\bibinfo{person}{Taslim Murad}, \bibinfo{person}{Sarwan Ali}, {and} \bibinfo{person}{Murray Patterson}.} \bibinfo{year}{2024}\natexlab{a}.
\newblock \showarticletitle{Weighted chaos game representation for molecular sequence classification}. In \bibinfo{booktitle}{\emph{Pacific-Asia Conference on Knowledge Discovery and Data Mining}}. Springer, \bibinfo{pages}{234--245}.
\newblock


\bibitem[Murad et~al\mbox{.}(2024c)]%
        {murad2024dance}
\bibfield{author}{\bibinfo{person}{Taslim Murad}, \bibinfo{person}{Prakash Chourasia}, \bibinfo{person}{Sarwan Ali}, \bibinfo{person}{Imdad~Ullah Khan}, {and} \bibinfo{person}{Murray Patterson}.} \bibinfo{year}{2024}\natexlab{c}.
\newblock \showarticletitle{Dance: Deep learning-assisted analysis of protein sequences using chaos enhanced kaleidoscopic images}.
\newblock \bibinfo{journal}{\emph{arXiv preprint arXiv:2409.06694}} (\bibinfo{year}{2024}).
\newblock


\bibitem[Murad et~al\mbox{.}(2024b)]%
        {murad2024advancing}
\bibfield{author}{\bibinfo{person}{Taslim Murad}, \bibinfo{person}{Prakash Chourasia}, \bibinfo{person}{Sarwan Ali}, {and} \bibinfo{person}{Murray Patterson}.} \bibinfo{year}{2024}\natexlab{b}.
\newblock \showarticletitle{Advancing protein-DNA binding site prediction: integrating sequence models and machine learning classifiers}. In \bibinfo{booktitle}{\emph{International Conference on Neural Information Processing}}. Springer, \bibinfo{pages}{356--371}.
\newblock


\bibitem[Rao et~al\mbox{.}(2019)]%
        {rao2019evaluating}
\bibfield{author}{\bibinfo{person}{Roshan Rao}, \bibinfo{person}{Nicholas Bhattacharya}, \bibinfo{person}{Neil Thomas}, \bibinfo{person}{Yan Duan}, \bibinfo{person}{Peter Chen}, \bibinfo{person}{John Canny}, \bibinfo{person}{Pieter Abbeel}, {and} \bibinfo{person}{Yun Song}.} \bibinfo{year}{2019}\natexlab{}.
\newblock \showarticletitle{Evaluating protein transfer learning with TAPE}.
\newblock \bibinfo{journal}{\emph{Advances in neural information processing systems}}  \bibinfo{volume}{32} (\bibinfo{year}{2019}).
\newblock


\bibitem[Tayebi et~al\mbox{.}(2023)]%
        {tayebi2023t}
\bibfield{author}{\bibinfo{person}{Zahra Tayebi}, \bibinfo{person}{Sarwan Ali}, \bibinfo{person}{Prakash Chourasia}, \bibinfo{person}{Taslim Murad}, {and} \bibinfo{person}{Murray Patterson}.} \bibinfo{year}{2023}\natexlab{}.
\newblock \showarticletitle{T Cell Receptor Protein Sequences and Sparse Coding: A Novel Approach to Cancer Classification}. In \bibinfo{booktitle}{\emph{International Conference on Neural Information Processing}}. Springer, \bibinfo{pages}{215--227}.
\newblock


\bibitem[Tayebi et~al\mbox{.}(2024)]%
        {tayebi2024pseaac2vec}
\bibfield{author}{\bibinfo{person}{Zahra Tayebi}, \bibinfo{person}{Sarwan Ali}, \bibinfo{person}{Taslim Murad}, \bibinfo{person}{Imdadullah Khan}, {and} \bibinfo{person}{Murray Patterson}.} \bibinfo{year}{2024}\natexlab{}.
\newblock \showarticletitle{PseAAC2Vec protein encoding for TCR protein sequence classification}.
\newblock \bibinfo{journal}{\emph{Computers in Biology and Medicine}}  \bibinfo{volume}{170} (\bibinfo{year}{2024}), \bibinfo{pages}{107956}.
\newblock


\bibitem[Tayebi et~al\mbox{.}(2021)]%
        {tayebi2021robust}
\bibfield{author}{\bibinfo{person}{Zahra Tayebi}, \bibinfo{person}{Sarwan Ali}, {and} \bibinfo{person}{Murray Patterson}.} \bibinfo{year}{2021}\natexlab{}.
\newblock \showarticletitle{Robust representation and efficient feature selection allows for effective clustering of sars-cov-2 variants}.
\newblock \bibinfo{journal}{\emph{Algorithms}} \bibinfo{volume}{14}, \bibinfo{number}{12} (\bibinfo{year}{2021}), \bibinfo{pages}{348}.
\newblock


\bibitem[Vaswani et~al\mbox{.}(2017)]%
        {vaswani2017attention}
\bibfield{author}{\bibinfo{person}{Ashish Vaswani}, \bibinfo{person}{Noam Shazeer}, \bibinfo{person}{Niki Parmar}, \bibinfo{person}{Jakob Uszkoreit}, \bibinfo{person}{Llion Jones}, \bibinfo{person}{Aidan~N Gomez}, \bibinfo{person}{{\L}ukasz Kaiser}, {and} \bibinfo{person}{Illia Polosukhin}.} \bibinfo{year}{2017}\natexlab{}.
\newblock \showarticletitle{Attention is all you need}. In \bibinfo{booktitle}{\emph{Advances in neural information processing systems}}. \bibinfo{pages}{5998--6008}.
\newblock


\bibitem[Zeng et~al\mbox{.}(2016)]%
        {zeng2016convolutional}
\bibfield{author}{\bibinfo{person}{H. Zeng}, \bibinfo{person}{M.D. Edwards}, \bibinfo{person}{G. Liu}, {and} \bibinfo{person}{D.K. Gifford}.} \bibinfo{year}{2016}\natexlab{}.
\newblock \showarticletitle{Convolutional neural network architectures for predicting DNA--protein binding}.
\newblock \bibinfo{journal}{\emph{Bioinformatics}} \bibinfo{volume}{32}, \bibinfo{number}{12} (\bibinfo{year}{2016}), \bibinfo{pages}{i121--i127}.
\newblock


\end{thebibliography}

\end{document}